\documentclass[sigconf]{acmart}
\AtBeginDocument{%
  \providecommand\BibTeX{{%
    \normalfont B\kern-0.5em{\scshape i\kern-0.25em b}\kern-0.8em\TeX}}}

\copyrightyear{2023}
\acmYear{2023}
\setcopyright{rightsretained}
\acmConference[MMAsia '23 Workshops]{ACM Multimedia Asia Workshops}{December 6--8, 2023}{Tainan, Taiwan}
\acmBooktitle{ACM Multimedia Asia Workshops (MMAsia '23 Workshops), December 6--8, 2023, Tainan, Taiwan}\acmDOI{10.1145/3611380.3630165}
\acmISBN{979-8-4007-0326-3/23/12}

\acmSubmissionID{355}

\begin{document}

\title{MAAIG:
Motion Analysis And Instruction Generation}

\author{Wei-Hsin Yeh}
\authornote{Both authors contributed equally to this research.}
\email{weihsinyeh168@gmail.com}
\orcid{0009-0001-2194-0419}
\affiliation{
  \institution{Academia Sinica}
  \city{Taipei}
  \country{Taiwan}
}

\author{Pei Hsin Lin}
\authornotemark[1]
\email{peihsin@caece.net}
\orcid{XXXX}
\affiliation{
  \institution{Academia Sinica}
  \city{Taipei}
  \country{Taiwan}
}

\author{Yu-An Su}
\authornotemark[1]
\orcid{0009-0004-9842-3251}
\email{yuansu@iis.sinica.edu.tw}
\affiliation{
  \institution{Academia Sinica}
  \city{Taipei}
  \country{Taiwan}
  \postcode{43017-6221}
}

\author{Wen Hsiang Cheng}
\authornotemark[1]
\orcid{0009-0000-6894-548X}
\email{bosszheng220@gmail.com}
\affiliation{
  \institution{Academia Sinica}
  \city{Taipei}
  \country{Taiwan}
}

\author{Lun-Wei Ku}
\authornotemark[1]
\orcid{0000-0003-2691-5404}
\email{lwku@iis.sinica.edu.tw}
\affiliation{
  \institution{Academia Sinica}
  \city{Taipei}
  \country{Taiwan}
}






\begin{abstract}
Many people engage in self-directed sports training at home but lack the real-time guidance of professional coaches, making them susceptible to injuries or the development of incorrect habits. In this paper, we propose a novel application framework called MAAIG(Motion Analysis And Instruction Generation). It can generate embedding vectors for each frame based on user-provided sports action videos. These embedding vectors are associated with the 3D skeleton of each frame and are further input into a pretrained T5 model. Ultimately, our model utilizes this information to generate specific sports instructions. It has the capability to identify potential issues and provide real-time guidance in a manner akin to professional coaches, helping users improve their sports skills and avoid injuries.
\end{abstract}

\ccsdesc[300]{Computing methodologies~Natural language generation}

\keywords{deep learning, instruction generation, nature language generating, computer vision}


\maketitle

\section{Introduction}
In this era, many individuals choose to self-learn various athletic movements at home due to constraints such as time limitations, inconvenient access to sports facilities, or other factors. However, autonomous learning of sports movements comes with several challenges, with one of the primary concerns being the absence of real-time guidance from professional coaches. In such circumstances, learners are susceptible to injuries or may develop incorrect exercise habits, potentially having adverse effects on their health and performance.

To address this issue, we propose a novel framework called Motion Analysis And Instruction Generation(MAAIG). This application accepts user-provided videos of sports movements and leverages advanced machine learning techniques to analyze motion characteristics and patterns within the videos. It can identify potential issues and provide real-time instructional language to users, akin to what a professional coach would say, aiding them in improving their athletic skills. This technology proves highly beneficial for accelerated learning and injury prevention, especially for those engaged in self-learning athletic activities at home.

Our approach is rooted in instruction generation. Initially, the videos captured by users are input into a pose recognition model to generate a dataset of 3D skeletal information. Subsequently, this dataset is transformed into embedding vectors, which are fed into the instruction generation model to provide guidance language to the users. Building upon this foundation, we introduce a novel architecture that redefines the instruction generation process. Traditional instruction generation often relies on static text or predefined categories, limiting its adaptability and real-time applicability. In contrast, our approach harnesses motion recognition from the source video, offering richer information to the instruction generation model.
\begin{figure*}[t]
  \includegraphics[width=\textwidth]{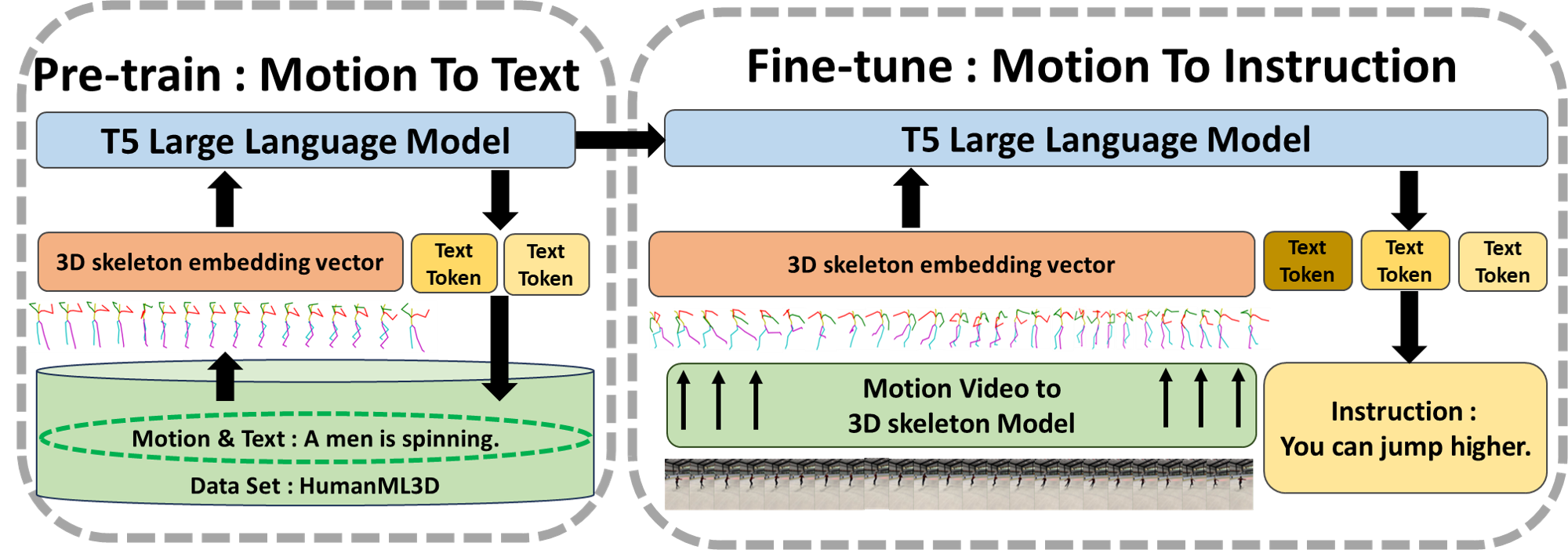}
  \caption{Framework of Motion To Instruction Model.}
  \Description{Pretrain on the task "Motion To Text" then Fine-tune on the task "Motion To Instruction"}
  \label{fig:teaser}
\end{figure*}
\section{Related Work}
\subsection{Motion to Text}
Several research studies on Motion2text \cite{guo2022tm2t} \cite{jiang2023motiongpt}.
In their work \cite{guo2022tm2t}, the author introduce a novel approach in which they transform a sequence of motions into a tokenized motion representation within a Code Book, employing the VQ-VAE \cite{DBLP:journals/corr/abs-1711-00937} framework. 
This process generates a compact yet semantically rich representation for 3-D motions. Leveraging quantified embedding vectors (namely motion tokens ), the authors employ a Transformer model to build a mapping from human motions to textual descriptions. This approach proves effective in connecting quantified motion tokens to quantified text tokens, enhancing the text generated by model can better describe the sequence of motion. 
However, it is worthy noting that the aforementioned research primarily focus on generating text which can capture the semantic essence of an entire motion sequence. When it comes to the task "motion to instruction", it is still too difficult to capture the detailed information of the sequence of motions.

In \cite{jiang2023motiongpt}, MotionGPT employs T5 as its underlying language model with motion-aware capabilities. The research involves two stage process: initial train models with pretraining and then instruction tuning. This approach results in improvements across various tasks and enhances the model's performance, particularly when faced with previously unseen tasks or prompts.

\subsection{Motion to Instruction}
In \cite{zhao20223d}, the framework for providing exercise feedback relies on Graph Convolutional Kernel. The author employee a classifier model to predict action labels (namely instruction labels). It is worthy of noting that instead of utilizing the 3D joint positions, the author take the DCT \cite{1672377} coefficients of joint trajectories as input of the classifier model. Furthermore, the predicted action labels are input to a feedback model, subsequently converted into the tensor. This resulting tensor is then fed into a correction model along with the DCT coefficients of joints. This mechanism enable the correction model to improve the accuracy of the corrected motion. This exercise feedback framework can generate the instruction label. We believe that the output text would be more valuable if it could provide precise instructions rather than just labels. In this way, the framework can offer more meaningful guidance for exercise routines.

\section{Methodology}
This paper presents a framework for converting skeletal information into language guidance, and the basic structure is depicted in Figure \ref{fig:teaser}. In the following sections, individual models will be discussed in detail.
\subsection{Human Pose Embedding}
The motion to instruction model takes skeleton-based data as input, which is obtained from pose estimation algorithms from videos proposed by HybrIK model \cite{DBLP:journals/corr/abs-2011-14672}, which estimates 3D pose via the twist-and-swing decomposition. Because our motion to instruction model is dedicated to assisting figure skaters in improving their athletic postures, and figure skating involves a multitude of airborne rotations and body twists, we choose the HybrIK \cite{DBLP:journals/corr/abs-2011-14672} model to assist us in constructing the skeletal structure from the video. 

Additionally, it's essential to note that figure skating takes place on an ice rink and our goal aims to provide accurate instruction. Therefore, we also endeavor to construct a 3-D skeleton to provide the motion to instruction model with distance-related information.

The input data consists of a sequence of frames, with each frame containing a set of 3-D pose joint coordinates. These coordinates are relative to the human body-part in the SMPL \cite{SMPL:2015} format with dimension ( 22x3 ). "22" means the number of joints and "3" means the x, y, z coordinates for each joints.

\subsection{Motion to Instruction Model}
Here we describe the proposed model in detail. The inputs to the model consist of 3-D skeleton-based data, with joint dimensions represented as  ( 22x3 ). Additionally, we apply linear transformation to convert our 3-D skeleton, which initially has a  dimension of 66, into 512-dimensional embedding. 

Our model is built upon the T5 (Text-to-Text Transfer Transformer) \cite{DBLP:journals/corr/abs-1910-10683} architecture introduced by Google Research in 2019, which represents a significant advancement in the field of natural language processing (NLP). T5 is a variant of the Transformer \cite{vaswani2023attention} architecture, notable for its pioneering self-attention mechanism that enables the model to process positional information across input sequences simultaneously. This capability is particularly valuable for capturing long-range dependencies in text. The fundamental idea behind T5 is to frame various NLP tasks as text-to-text transformations. It adopts an end-to-end approach where both inputs and outputs are treated as text sequences. This implies that tasks spanning question answering, text classification, text generation, summarization, translation, and more can all be cast as processes that map one text sequence to another. T5's architectural design comprises two essential components: the encoder and the decoder, which are the core constituents of the Transformer architecture. The encoder is responsible for encoding input text into an intermediate representation, while the decoder deciphers this intermediate representation into the desired output text. During the training process, T5 learns its parameters by minimizing the disparity between predicted output text and the actual target text. This enables the model to generate high-quality outputs during inference. Whether the task at hand involves answering questions, classifying text, generating text, summarizing, or translating, T5's text-to-text framework offers a unified approach. This versatility empowers T5 as a robust and adaptable NLP model, obviating the necessity to design distinct models for individual tasks. Instead, it requires only adjustments to input and output text specifications.

To tokenize instructions, we used the Tokenizer from mT5 (multilingual Text-to-Text Transfer Transformer). mT5 is a multi-language NLP model based on the Transformer architecture, known for its excellent cross-language processing capabilities. This tokenizer facilitates the tokenization, processing, and representation of textual instructions in a manner compatible with our model's architecture. Thus, the model can simultaneously handle 3D skeletal data and textual instructions to perform the required tasks.

Furthermore, to equip the model with the capability to handle 3-D skeletal data and generate language, we conducted pretraining using the HumanML3D dataset \cite{Plappert2016} through T5. HumanML3D is a pretrained three-dimensional human pose recognition model designed to integrate three-dimensional skeletal data with tasks related to human actions. This model focuses on understanding and interpreting three-dimensional human poses, which hold significant value in various applications such as sports analysis, motion capture, human-computer interaction, and virtual reality. Key features and functionalities of HumanML3D include processing three-dimensional skeletal data, pretrained representations, multi-task learning, and real-time performance. It can accept three-dimensional skeletal data captured from deep learning cameras, sensors, or other devices as input, identify and track the positions, orientations, and movements of human joints, and reconstruct three-dimensional human poses. Through large-scale pretraining, HumanML3D can acquire a universal representation of human poses applicable to various tasks and datasets.

\begin{figure}[h]
  \centering
  \includegraphics[width=\linewidth]{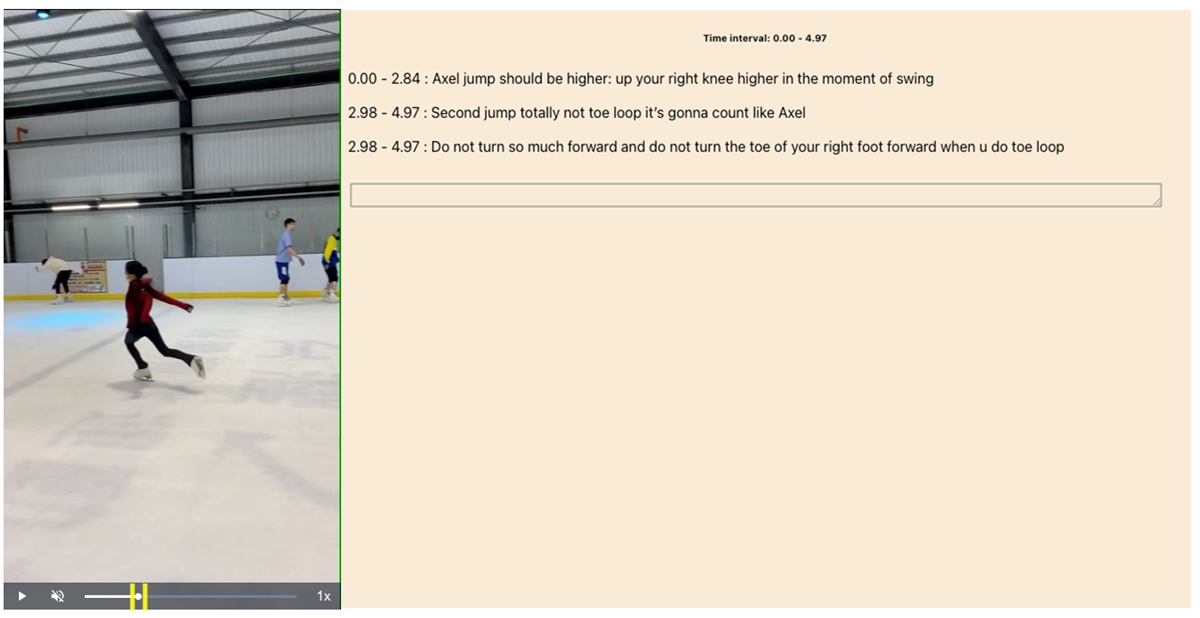}
  \caption{The annotation instruction system. The coach will first select the video interval by the yellow bars at the left, then enter the corresponding instruction or feedback on the right. }
  \Description{The user interface of the annotating system.}
  \label{fig:system}
\end{figure}

\section{Experiments}
In this section, we commence by providing a detailed description of the MAAIG, a private dataset that we collected specifically for the use in this experiment. Following this, we will delve into the intricacies of our framework implementations. Finally, we will present empirical results, showcasing the efficacy of our framework's instruction generation capabilities, assessed through a diverse array of machine translation evaluation metrics, namely Bleu \cite{papineni-etal-2002-bleu}, METEOR \cite{banerjee-lavie-2005-meteor}, and ROGUE \cite{grusky-2023-rogue}.

\subsection{YourSkatingCoach Dataset}
Given the lack of existing motion and instruction pair dataset, We collect our own figure skating dataset for this specific task by our self-developed annotation instruction system. Experts can select specific time segments from the database's videos and annotate them with instruction. An example of the selected motion clip and its corresponding instructions are shown in Figure \ref{fig:system}. In this section, we outline the process for acquiring the YourSkatingCoach, which involves three distinct phases: Motion Collection, Instruction Collection, and Dataset Generation. 

\textbf{Motion Collection.}  
We obtain video footage featuring two skaters performing the Axel jump, a maneuver in skating that involves transitioning from the forward outside edge of one skate to the backward outside edge of the other, encompassing one (or more) and a half turns in mid-air. Subsequently, we categorized these videos into two distinct groups: "Axel" for single jump performances and "Axel\_com" (Axel combo) for sequences comprising one or more consecutive jumps. Nevertheless, due to the inadequate collection of data in the Axel dataset, we have opted to exclusively utilize the Axel\_com subset in this experiment. At first, there were 89 videos of athletes performing these two motions.

\textbf{Instruction Collection.}
To acquire the instructions for the motions we collected, we hire a professional skater to annotate the video tapes we collected. The annotation process works as follows: firstly, the skater watch the video taken from the last phase and select a "start time" and an "end time" of the clip and leave an instruction on how the performers can improve their Axel moves. In the end, we clip out 164 motion clips from the original 89 videos, and divide them to 90/10 for training and testing.

\textbf{Dataset Generation.}
With the motion dataset obtained from the Motion Collection and the instruction dataset acquired through the Instruction Collection, we can now construct a comprehensive dataset composed of pairs comprising motion clips and their respective instructions. This process involves segmenting the videos based on the "start time" and "end time" chosen by the annotator, followed by merging them with the corresponding annotations provided by the annotator. In scenarios where multiple instructions are assigned to the same time interval within a single video, we unify these instructions into a coherent directive by connecting them using a designated separator symbol. 

\begin{table*}[t]
    \centering
    \caption{Evaluation Metrics for Different Models}
    \begin{tabular}{lllccccccc}
        \toprule
        \textbf{Model} & \textbf{Pretrain} & \textbf{Bleu\_1} & \textbf{Bleu\_2} & \textbf{Bleu\_3} & \textbf{Bleu\_4} & \textbf{METEOR} & \textbf{ROUGE\_L} \\
        \midrule
        Transformer & N/A & 0.100418 & 0.067011 & 0.047725 & 0.037824 & 0.084673 & 0.180627 \\
        Transformer & HumanML3D (world) & 0.250667 & 0.170773 & 0.133011 & 0.110372 & 0.161843 & 0.212154 \\
        Transformer & HumanML3D (local) & 0.268576 & 0.163972 & 0.106748 & 0.076333 & 0.125742 & 0.202856 \\
        T5 & N/A & 0.371186 & 0.284594 & 0.235283 & 0.204323 & 0.20255 & 0.329223 \\
        T5 & HumanML3D (world) & 0.427747 & 0.335629 & 0.282911 & 0.250055 & 0.216211 & 0.380814 \\
        T5 & HumanML3D (local) & \textbf{0.439764} & \textbf{0.345505} & \textbf{0.299627} & \textbf{0.271761} & \textbf{0.220812} & \textbf{0.388061} \\
        \bottomrule
    \end{tabular}
    \label{tab:evaluation-metrics}
\end{table*}

\subsection{Implementation Details}
Initially, we embark on model development from the ground up using T5 architecture. However, this approach yields subpar results owing to the inherent limitations of our YourSkatingCoach Dataset, which comprises a relatively modest collection of just 164 video clips. The constraints imposed by this dataset made it unfeasible to train a sufficiently compact Large Language Model capable of processing textual content effectively.

In response to these constraints, we turn to the HumanML3D dataset \cite{petrovich2022temos}, a valuable resource encompassing a substantial repository of 44,970 textual descriptions corresponding to 14,616 3D human motions. This dataset serves as a pivotal pretraining source for our model, empowering it to attain a holistic comprehension of motion representations.

However, it has come to our attention that there exists a difference in the skeletal data between YourSkatingCoach and HumanML3D. Specifically, the data in YourSkatingCoach is represented in a local coordinate system, while HumanML3D utilizes the world coordinate system. As a result, we have explored an alternative setting where we remap HumanML3D into the local coordinate system to address this discrepancy.

In addition to our exploration of the T5 architecture, we extended our experimentation to include the Transformer architecture. Our experiments cover the following six distinct settings:
\begin{enumerate}
  \item Train from scratch using Transformer.
  \item Fine-tune from HumanML3D pretrain in world coordinate system using Transformer.
  \item Fine-tune from HumanML3D pretrain in local coordinate system using Transformer.
  \item Train from scratch using T5.
  \item Fine-tune from HumanML3D pretrain in world coordinate system using T5.
  \item Fine-tune from HumanML3D pretrain in local coordinate system using T5.
\end{enumerate}

\subsection{Results}
In Table \ref{tab:evaluation-metrics}, we present the results of our experiment. Two key observations emerge from this table: (1) Pretraining on a large dataset helps mitigate the limitations of a small dataset. (2) Converting the coordinate systems further enhances performance.

\textbf{Comparing With or Without Pretraining.}
The table clearly illustrates the substantial performance improvement achieved by using pretraining across both the Transformer and T5 architectures. This demonstrates the efficacy of pretraining on a large dataset in overcoming the limitations of a smaller dataset, thereby enhancing the language understanding within our model. Notably, all scores exhibit a significant elevation when HumanML3D is utilized as the pretraining source.

\textbf{Comparing With or Without Converting Coordination.}
The table reveals that performance is notably enhanced when employing the HumanML3D dataset in the local coordinate system compared to the world coordinate system. This observation underscores the critical role of data compatibility in transferring language understanding from one dataset to another. Remarkably, when YourSkatingCoach and HumanML3D share the same coordinate system, they yield the optimal result in the T5 architectures in the setting of Fine-tune from HumanML3D pretrain in local coordinate system.
\begin{figure}
  \centering
  \includegraphics[width=\linewidth]{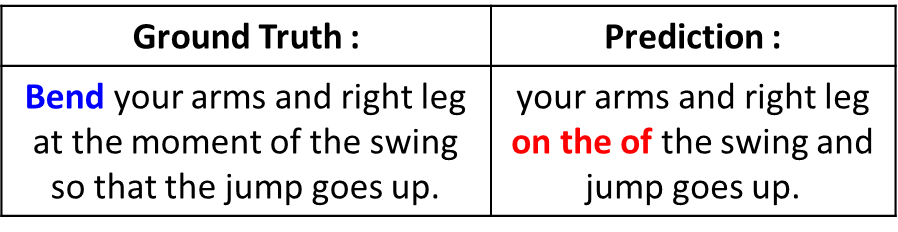}
  \caption{Example of failing to generate complete sentence. }
  \Description{A video and its instruction labels.}
  \label{fig:prediction_failed}
\end{figure}

\subsection{Discussion}
While pretraining on the large dataset with T5 significantly boost the performance, there were still parts where the model performed poorly. For example we observe that the model cannot distinguish left from right and vice versa. We suspect that this can be attributed to the scarcity of the annotation regarding left and right. Secondly, the vanilla transformer model have failed to generate fluent sentences, some of the common failures are: omitting verbs, repetitive words, and violating grammar, as shown in Figure \ref{fig:prediction_failed}. Though these issues were all largely mitigated when we switch to the T5 implementation, each evaluating score still remains rooms to improve. 

In the future, we aim to initially convert the video actions into world coordinates system by estimating implicit and explicit camera parameters to let model learn left and right. After converting human pose from camera coordinate system to world coordinate system, we will train our "MAAIG" model in world coordinate system, enabling it to learn the direction-related information. Subsequently, we can also take the kinematic information of skeleton in world coordinate system, e.g.: angular velocity, linear velocity, and rotation as the input of "MMAIG" model. Nonetheless, we will employ Part-of-speech (POS) tagging on the instruction, which can ensure the output of our model is grammarly correct. Also, it is non-trivial to study the effect of different movement speed in terms of the model performance, we plan to set off a variety of experiments to ensure our model's stability under arbitrary situations.

\section{Conclusion}
In this paper, we introduce a model based on the T5 architecture for generating sports coaching instructions. This model utilizes pose recognition technology to generate 3D skeletons, which are then transformed into embedding vectors. These embedding vectors are fed into the instruction generation model to produce coaching instructions. In our experiments, we applied this model to the "Axel combination" movement in the context of ice skating. By incorporating a pretrained HumanML3D model and using BLEU scores as the measure of accuracy, we achieved a BLEU score of 0.439. This indicates that language models are capable of reading and analyzing input 3D skeleton data, resulting in the generation of corresponding coaching instructions. Our model provides a promising tool for sports coaches and athletes, as it can generate personalized instructions based on the user's actual movements. It has the potential to play a significant role in various sports training and sports science applications.

\begin{acks}
 We thank figure skating coach Kristina Stepanova for helping with video shooting and instruction generation. This work is supported by the National Science and Technology Council of Taiwan under grants 112-2425-H-007-002- and by the Academia Sinica and National Tsing Hua University collaborative project.
\end{acks}

\bibliographystyle{ACM-Reference-Format}
\bibliography{sample-authordraft}


\begin{thebibliography}{14}


\ifx \showCODEN    \undefined \def \showCODEN     #1{\unskip}     \fi
\ifx \showDOI      \undefined \def \showDOI       #1{#1}\fi
\ifx \showISBNx    \undefined \def \showISBNx     #1{\unskip}     \fi
\ifx \showISBNxiii \undefined \def \showISBNxiii  #1{\unskip}     \fi
\ifx \showISSN     \undefined \def \showISSN      #1{\unskip}     \fi
\ifx \showLCCN     \undefined \def \showLCCN      #1{\unskip}     \fi
\ifx \shownote     \undefined \def \shownote      #1{#1}          \fi
\ifx \showarticletitle \undefined \def \showarticletitle #1{#1}   \fi
\ifx \showURL      \undefined \def \showURL       {\relax}        \fi
\providecommand\bibfield[2]{#2}
\providecommand\bibinfo[2]{#2}
\providecommand\natexlab[1]{#1}
\providecommand\showeprint[2][]{arXiv:#2}

\bibitem[Ahmed et~al\mbox{.}(1974)]%
        {1672377}
\bibfield{author}{\bibinfo{person}{N. Ahmed}, \bibinfo{person}{T. Natarajan}, {and} \bibinfo{person}{K.R. Rao}.} \bibinfo{year}{1974}\natexlab{}.
\newblock \showarticletitle{Discrete Cosine Transform}.
\newblock \bibinfo{journal}{\emph{IEEE Trans. Comput.}} \bibinfo{volume}{C-23}, \bibinfo{number}{1} (\bibinfo{year}{1974}), \bibinfo{pages}{90--93}.
\newblock
\urldef\tempurl%
\url{https://doi.org/10.1109/T-C.1974.223784}
\showDOI{\tempurl}


\bibitem[Banerjee and Lavie(2005)]%
        {banerjee-lavie-2005-meteor}
\bibfield{author}{\bibinfo{person}{Satanjeev Banerjee} {and} \bibinfo{person}{Alon Lavie}.} \bibinfo{year}{2005}\natexlab{}.
\newblock \showarticletitle{{METEOR}: An Automatic Metric for {MT} Evaluation with Improved Correlation with Human Judgments}. In \bibinfo{booktitle}{\emph{Proceedings of the {ACL} Workshop on Intrinsic and Extrinsic Evaluation Measures for Machine Translation and/or Summarization}}. \bibinfo{publisher}{Association for Computational Linguistics}, \bibinfo{address}{Ann Arbor, Michigan}, \bibinfo{pages}{65--72}.
\newblock
\urldef\tempurl%
\url{https://aclanthology.org/W05-0909}
\showURL{%
\tempurl}


\bibitem[Grusky(2023)]%
        {grusky-2023-rogue}
\bibfield{author}{\bibinfo{person}{Max Grusky}.} \bibinfo{year}{2023}\natexlab{}.
\newblock \showarticletitle{Rogue Scores}. In \bibinfo{booktitle}{\emph{Proceedings of the 61st Annual Meeting of the Association for Computational Linguistics (Volume 1: Long Papers)}}. \bibinfo{publisher}{Association for Computational Linguistics}, \bibinfo{address}{Toronto, Canada}, \bibinfo{pages}{1914--1934}.
\newblock
\urldef\tempurl%
\url{https://doi.org/10.18653/v1/2023.acl-long.107}
\showDOI{\tempurl}


\bibitem[Guo et~al\mbox{.}(2022)]%
        {guo2022tm2t}
\bibfield{author}{\bibinfo{person}{Chuan Guo}, \bibinfo{person}{Xinxin Zuo}, \bibinfo{person}{Sen Wang}, {and} \bibinfo{person}{Li Cheng}.} \bibinfo{year}{2022}\natexlab{}.
\newblock \bibinfo{title}{TM2T: Stochastic and Tokenized Modeling for the Reciprocal Generation of 3D Human Motions and Texts}.
\newblock
\newblock
\showeprint[arxiv]{2207.01696}~[cs.CV]


\bibitem[Jiang et~al\mbox{.}(2023)]%
        {jiang2023motiongpt}
\bibfield{author}{\bibinfo{person}{Biao Jiang}, \bibinfo{person}{Xin Chen}, \bibinfo{person}{Wen Liu}, \bibinfo{person}{Jingyi Yu}, \bibinfo{person}{Gang Yu}, {and} \bibinfo{person}{Tao Chen}.} \bibinfo{year}{2023}\natexlab{}.
\newblock \bibinfo{title}{MotionGPT: Human Motion as a Foreign Language}.
\newblock
\newblock
\showeprint[arxiv]{2306.14795}~[cs.CV]


\bibitem[Li et~al\mbox{.}(2020)]%
        {DBLP:journals/corr/abs-2011-14672}
\bibfield{author}{\bibinfo{person}{Jiefeng Li}, \bibinfo{person}{Chao Xu}, \bibinfo{person}{Zhicun Chen}, \bibinfo{person}{Siyuan Bian}, \bibinfo{person}{Lixin Yang}, {and} \bibinfo{person}{Cewu Lu}.} \bibinfo{year}{2020}\natexlab{}.
\newblock \showarticletitle{HybrIK: {A} Hybrid Analytical-Neural Inverse Kinematics Solution for 3D Human Pose and Shape Estimation}.
\newblock \bibinfo{journal}{\emph{CoRR}}  \bibinfo{volume}{abs/2011.14672} (\bibinfo{year}{2020}).
\newblock
\showeprint[arXiv]{2011.14672}
\urldef\tempurl%
\url{https://arxiv.org/abs/2011.14672}
\showURL{%
\tempurl}


\bibitem[Loper et~al\mbox{.}(2015)]%
        {SMPL:2015}
\bibfield{author}{\bibinfo{person}{Matthew Loper}, \bibinfo{person}{Naureen Mahmood}, \bibinfo{person}{Javier Romero}, \bibinfo{person}{Gerard Pons-Moll}, {and} \bibinfo{person}{Michael~J. Black}.} \bibinfo{year}{2015}\natexlab{}.
\newblock \showarticletitle{{SMPL}: A Skinned Multi-Person Linear Model}.
\newblock \bibinfo{journal}{\emph{ACM Trans. Graphics (Proc. SIGGRAPH Asia)}} \bibinfo{volume}{34}, \bibinfo{number}{6} (\bibinfo{date}{Oct.} \bibinfo{year}{2015}), \bibinfo{pages}{248:1--248:16}.
\newblock
\urldef\tempurl%
\url{https://doi.org/10.1145/2816795.2818013}
\showDOI{\tempurl}


\bibitem[Papineni et~al\mbox{.}(2002)]%
        {papineni-etal-2002-bleu}
\bibfield{author}{\bibinfo{person}{Kishore Papineni}, \bibinfo{person}{Salim Roukos}, \bibinfo{person}{Todd Ward}, {and} \bibinfo{person}{Wei-Jing Zhu}.} \bibinfo{year}{2002}\natexlab{}.
\newblock \showarticletitle{{B}leu: a Method for Automatic Evaluation of Machine Translation}. In \bibinfo{booktitle}{\emph{Proceedings of the 40th Annual Meeting of the Association for Computational Linguistics}}. \bibinfo{publisher}{Association for Computational Linguistics}, \bibinfo{address}{Philadelphia, Pennsylvania, USA}, \bibinfo{pages}{311--318}.
\newblock
\urldef\tempurl%
\url{https://doi.org/10.3115/1073083.1073135}
\showDOI{\tempurl}


\bibitem[Petrovich et~al\mbox{.}(2022)]%
        {petrovich2022temos}
\bibfield{author}{\bibinfo{person}{Mathis Petrovich}, \bibinfo{person}{Michael~J. Black}, {and} \bibinfo{person}{Gül Varol}.} \bibinfo{year}{2022}\natexlab{}.
\newblock \bibinfo{title}{TEMOS: Generating diverse human motions from textual descriptions}.
\newblock
\newblock
\showeprint[arxiv]{2204.14109}~[cs.CV]


\bibitem[Plappert et~al\mbox{.}(2016)]%
        {Plappert2016}
\bibfield{author}{\bibinfo{person}{Matthias Plappert}, \bibinfo{person}{Christian Mandery}, {and} \bibinfo{person}{Tamim Asfour}.} \bibinfo{year}{2016}\natexlab{}.
\newblock \showarticletitle{The {KIT} Motion-Language Dataset}.
\newblock \bibinfo{journal}{\emph{Big Data}} \bibinfo{volume}{4}, \bibinfo{number}{4} (\bibinfo{date}{dec} \bibinfo{year}{2016}), \bibinfo{pages}{236--252}.
\newblock
\urldef\tempurl%
\url{https://doi.org/10.1089/big.2016.0028}
\showDOI{\tempurl}


\bibitem[Raffel et~al\mbox{.}(2019)]%
        {DBLP:journals/corr/abs-1910-10683}
\bibfield{author}{\bibinfo{person}{Colin Raffel}, \bibinfo{person}{Noam Shazeer}, \bibinfo{person}{Adam Roberts}, \bibinfo{person}{Katherine Lee}, \bibinfo{person}{Sharan Narang}, \bibinfo{person}{Michael Matena}, \bibinfo{person}{Yanqi Zhou}, \bibinfo{person}{Wei Li}, {and} \bibinfo{person}{Peter~J. Liu}.} \bibinfo{year}{2019}\natexlab{}.
\newblock \showarticletitle{Exploring the Limits of Transfer Learning with a Unified Text-to-Text Transformer}.
\newblock \bibinfo{journal}{\emph{CoRR}}  \bibinfo{volume}{abs/1910.10683} (\bibinfo{year}{2019}).
\newblock
\showeprint[arXiv]{1910.10683}
\urldef\tempurl%
\url{http://arxiv.org/abs/1910.10683}
\showURL{%
\tempurl}


\bibitem[van~den Oord et~al\mbox{.}(2017)]%
        {DBLP:journals/corr/abs-1711-00937}
\bibfield{author}{\bibinfo{person}{A{\"{a}}ron van~den Oord}, \bibinfo{person}{Oriol Vinyals}, {and} \bibinfo{person}{Koray Kavukcuoglu}.} \bibinfo{year}{2017}\natexlab{}.
\newblock \showarticletitle{Neural Discrete Representation Learning}.
\newblock \bibinfo{journal}{\emph{CoRR}}  \bibinfo{volume}{abs/1711.00937} (\bibinfo{year}{2017}).
\newblock
\showeprint[arXiv]{1711.00937}
\urldef\tempurl%
\url{http://arxiv.org/abs/1711.00937}
\showURL{%
\tempurl}


\bibitem[Vaswani et~al\mbox{.}(2023)]%
        {vaswani2023attention}
\bibfield{author}{\bibinfo{person}{Ashish Vaswani}, \bibinfo{person}{Noam Shazeer}, \bibinfo{person}{Niki Parmar}, \bibinfo{person}{Jakob Uszkoreit}, \bibinfo{person}{Llion Jones}, \bibinfo{person}{Aidan~N. Gomez}, \bibinfo{person}{Lukasz Kaiser}, {and} \bibinfo{person}{Illia Polosukhin}.} \bibinfo{year}{2023}\natexlab{}.
\newblock \bibinfo{title}{Attention Is All You Need}.
\newblock
\newblock
\showeprint[arxiv]{1706.03762}~[cs.CL]


\bibitem[Zhao et~al\mbox{.}(2022)]%
        {zhao20223d}
\bibfield{author}{\bibinfo{person}{Ziyi Zhao}, \bibinfo{person}{Sena Kiciroglu}, \bibinfo{person}{Hugues Vinzant}, \bibinfo{person}{Yuan Cheng}, \bibinfo{person}{Isinsu Katircioglu}, \bibinfo{person}{Mathieu Salzmann}, {and} \bibinfo{person}{Pascal Fua}.} \bibinfo{year}{2022}\natexlab{}.
\newblock \bibinfo{title}{3D Pose Based Feedback for Physical Exercises}.
\newblock
\newblock
\showeprint[arxiv]{2208.03257}~[cs.CV]


\end{thebibliography}

\appendix

\end{document}